\documentclass[letterpaper,journal]{IEEEtran}
\usepackage{amsmath,amsfonts,amssymb}
\usepackage{array}
\usepackage{graphicx}
\usepackage{booktabs}
\usepackage{adjustbox}
\usepackage{tabularx}
\newcolumntype{Y}{>{\raggedright\arraybackslash}X}
\usepackage{xurl}
\usepackage{stfloats}
\usepackage{url}
\usepackage{cite}

\graphicspath{{./}{pic/}}

\newcommand{\pihimtablecaption}[2][]{%
\caption{#2}%
\if\relax\detokenize{#1}\relax\else\label{#1}\fi
}

\begin{document}
\renewcommand{\topfraction}{0.95}
\renewcommand{\bottomfraction}{0.85}
\renewcommand{\textfraction}{0.05}
\renewcommand{\floatpagefraction}{0.80}
\setcounter{topnumber}{4}
\setcounter{bottomnumber}{2}
\setcounter{totalnumber}{6}

\title{Physics-Knowledge-Guided Hybrid Neural Learning for Arctic Sea Ice Concentration Evolution and Short-Range Prediction}

\author{Maqun Zhang$^1$, Feng Gao$^1$, Wankun Chen$^1$, Hui Yu$^2$, Yanhai Gan$^{1,*}$, and Junyu Dong$^{1,*}$%
\thanks{This work was supported by the Leverhulme Trust through Project VP1-2020-044; the National Natural Science Foundation of China under Grant 42406192; the Fundamental Research Funds for the Central Universities under Grants 202413040 and 202572015; the National Science and Technology Major Project of China under Grant 2022ZD0117201; and the Postdoctoral Project of Qingdao under Grant QDBSH20240102021.}%
\thanks{Corresponding authors: Yanhai Gan (ganyanhai@ouc.edu.cn) and Junyu Dong (dongjunyu@ouc.edu.cn).}%
\thanks{Maqun Zhang, Feng Gao, Wankun Chen, Yanhai Gan, and Junyu Dong are with the State Key Laboratory of Physical Oceanography and Faculty of Information Science and Engineering, Ocean University of China, Qingdao 266100, China. Hui Yu is with the School of Psychology and Neuroscience, University of Glasgow, Glasgow G12 8QQ, U.K.}}

\maketitle

\begin{abstract}
Accurate modeling of sea ice concentration (SIC) evolution is essential for polar climate assessment and short-range sea ice prediction. Numerical and data-driven approaches constitute major foundations for SIC modeling, but the former often require complex parameterizations and substantial computation, whereas the latter rarely encode physical dependencies explicitly. This study presents the Physics-Informed Hybrid Ice Model (PIHIM), a differentiable data-driven hybrid ice model for daily SIC evolution that organizes its network structure according to the physical dependencies encoded in the sea ice continuity equation and explicitly accounts for dynamical transport, thermodynamically driven areal growth and loss, and unresolved local processes. PIHIM preserves the representation capacity of deep learning while providing a process-decomposed formulation of ice displacement, freeze-melt areal change, and local error closure. Two evaluation settings are adopted: reanalysis-forced simulation examines SIC evolution stability under reanalysis forcing, and forecast-forced prediction assesses short-range performance under forecast-forced conditions, with reanalysis and observational SIC serving as verification references. Results indicate enhanced ice-edge preservation and error-growth control in reanalysis-forced simulation, while PIHIM retains measurable short-range prediction skill under forecast-forced conditions. Our code will be made publicly available after the paper is accepted.
\end{abstract}


\section{Introduction}

\IEEEPARstart{P}{olar} sea ice constitutes a key component of the climate system, modulating air-sea heat exchange, ocean circulation, polar ecosystems, and Arctic navigation \cite{ref1,ref2,ref3,ref4,ref5,ref8}. Accurate modeling of SIC evolution is therefore essential for climate monitoring and polar operations. However, SIC changes are jointly governed by atmospheric forcing, oceanic conditions, sea ice transport, and thermodynamic phase changes \cite{ref9,ref10,ref11,ref12}. Their nonlinear, seasonal, and spatially heterogeneous nature complicates characterization of ice-edge displacement and low-concentration sea ice in the marginal ice zone and during freeze/melt transitions \cite{ref13,ref14}.

Existing SIC evolution modeling approaches can be broadly grouped into physics-based numerical models and data-driven models \cite{ref15,ref16,ref17,ref18}. Numerical models explicitly represent sea ice dynamics and thermodynamics, but their performance depends on parameterizations of unresolved processes, and high-resolution simulations remain computationally demanding \cite{ref9,ref10,ref15,ref19}. Deep learning has also been widely used for spatiotemporal SIC prediction \cite{ref16,ref17,ref18,ref21}. Nevertheless, purely data-driven approaches commonly formulate SIC evolution as an end-to-end mapping, with internal states that do not readily correspond to dynamical transport, thermodynamic areal growth and decay, or residual closure. This opacity complicates error attribution and may weaken interpretability in autoregressive integration, under distribution shifts, and at high-gradient ice edges \cite{ref22,ref23,ref24}.

These considerations motivate a modeling strategy that combines process organization with learnable correction. 
The continuity equation separates horizontal transport from source-sink changes \cite{ref9,ref10,ref27,ref28}, while learnable residual closure compensates for simplified physics, numerical discretization, and unresolved processes \cite{ref22,ref26,ref29,ref30}.
Evaluation should also distinguish two settings: reanalysis-forced simulation, which examines whether the model can stably evolve SIC under prescribed reanalysis forcings, and forecast-forced prediction, which assesses short-range performance when boundary conditions are supplied by forecast products \cite{ref15,ref19,ref21,ref31}. These considerations are summarized schematically in Fig. 1.

\begin{figure}
	\centering
	\includegraphics[width= 1\columnwidth]{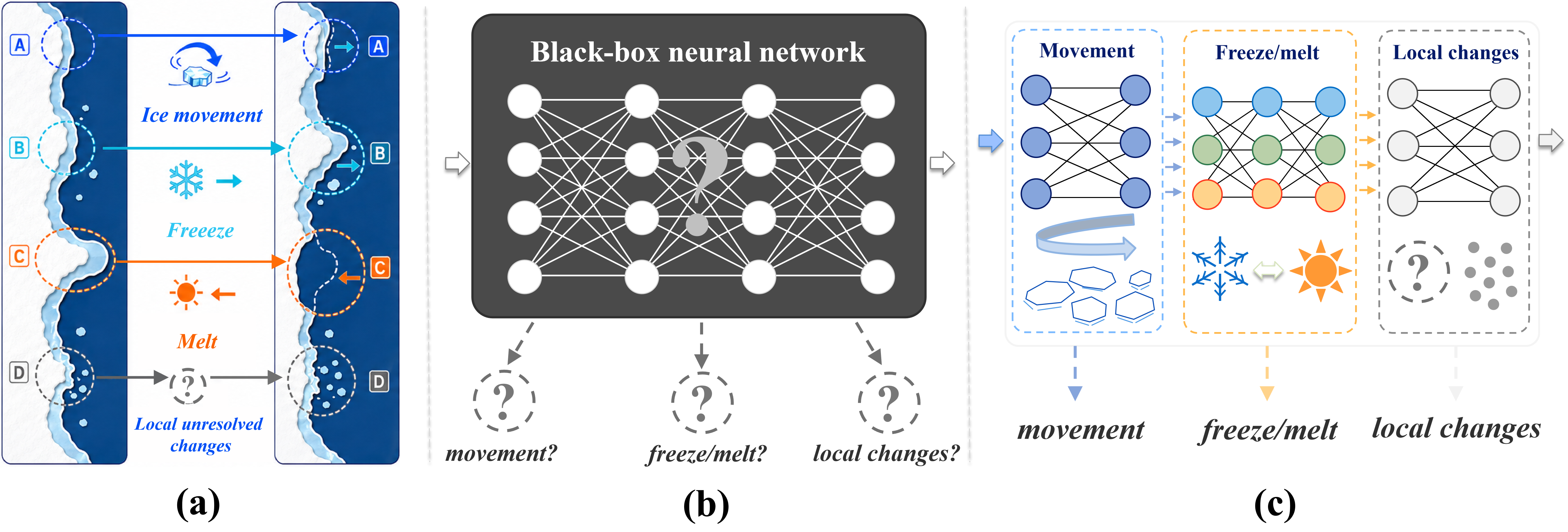}
	\caption{Conceptual overview. (a) SIC evolution is jointly driven by ice movement, freeze/melt processes, and local unresolved changes. (b) Black-box neural networks cannot explicitly attribute SIC changes to different physical processes. (c) PIHIM decomposes neural prediction into process-guided components for movement, freeze/melt, and local changes.}
	\label{fig:intro}
\end{figure}

To this end, this study proposes PIHIM, a differentiable hybrid framework for daily SIC evolution guided by the sea ice continuity equation. PIHIM decomposes one-step evolution into dynamical transport, a lightweight thermodynamics-guided source-term representation, and residual compensation. The dynamical module represents ice-motion-driven redistribution, the thermodynamic module constrains freeze/melt branches through sign constraints and SIC-based area-fraction weighting to describe areal change, and a lightweight residual-closure network compensates for local errors arising from simplified physics, discretization, and unresolved processes.

The main contributions are as follows:

\begin{enumerate}

\item We develop a differentiable data-driven hybrid ice model for daily SIC evolution, organizing its network structure according to the physical dependencies encoded in the sea ice continuity equation.

\item We introduce a lightweight, data-driven thermodynamics-guided source-term representation that avoids the computational burden of a full thermodynamic sea ice model, guides learning of SIC areal change through area-fraction-weighted freeze/melt branches, and supports analysis of seasonal thermodynamic responses.

\item We establish a two-setting evaluation framework that distinguishes reanalysis-forced simulation from forecast-forced prediction, separately assessing evolution stability and module contributions under reanalysis forcing, and short-range prediction performance under forecast-forced conditions.

\end{enumerate}

\vspace{1em} 

\section{Related Work}

\begin{figure*}[!t]

\centering

\textbf{(a)}\\[-0.2em]
\includegraphics[width=0.82\textwidth]{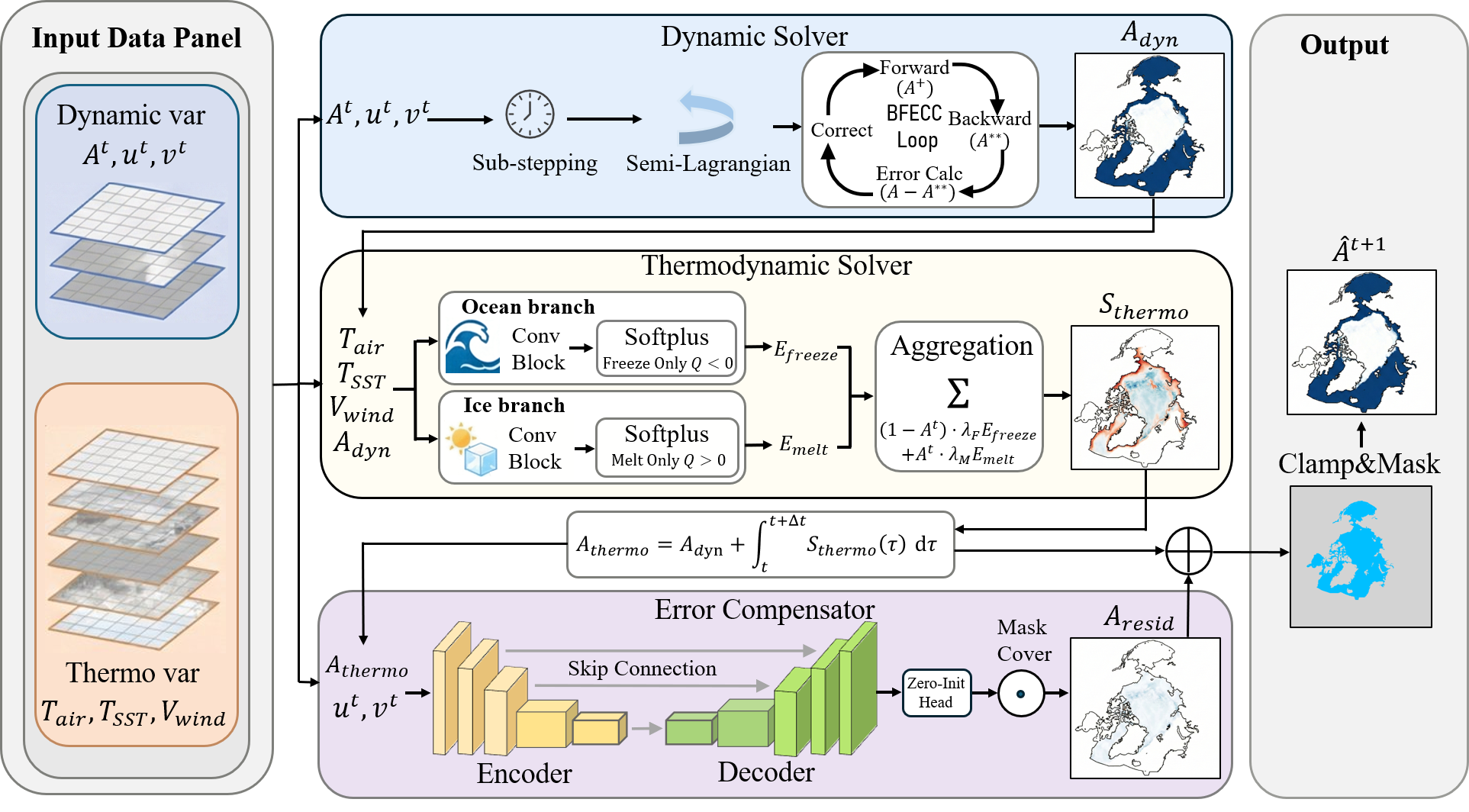}

\vspace{0.35em}

\textbf{(b)}\\[-0.2em]
\includegraphics[width=0.8\textwidth]{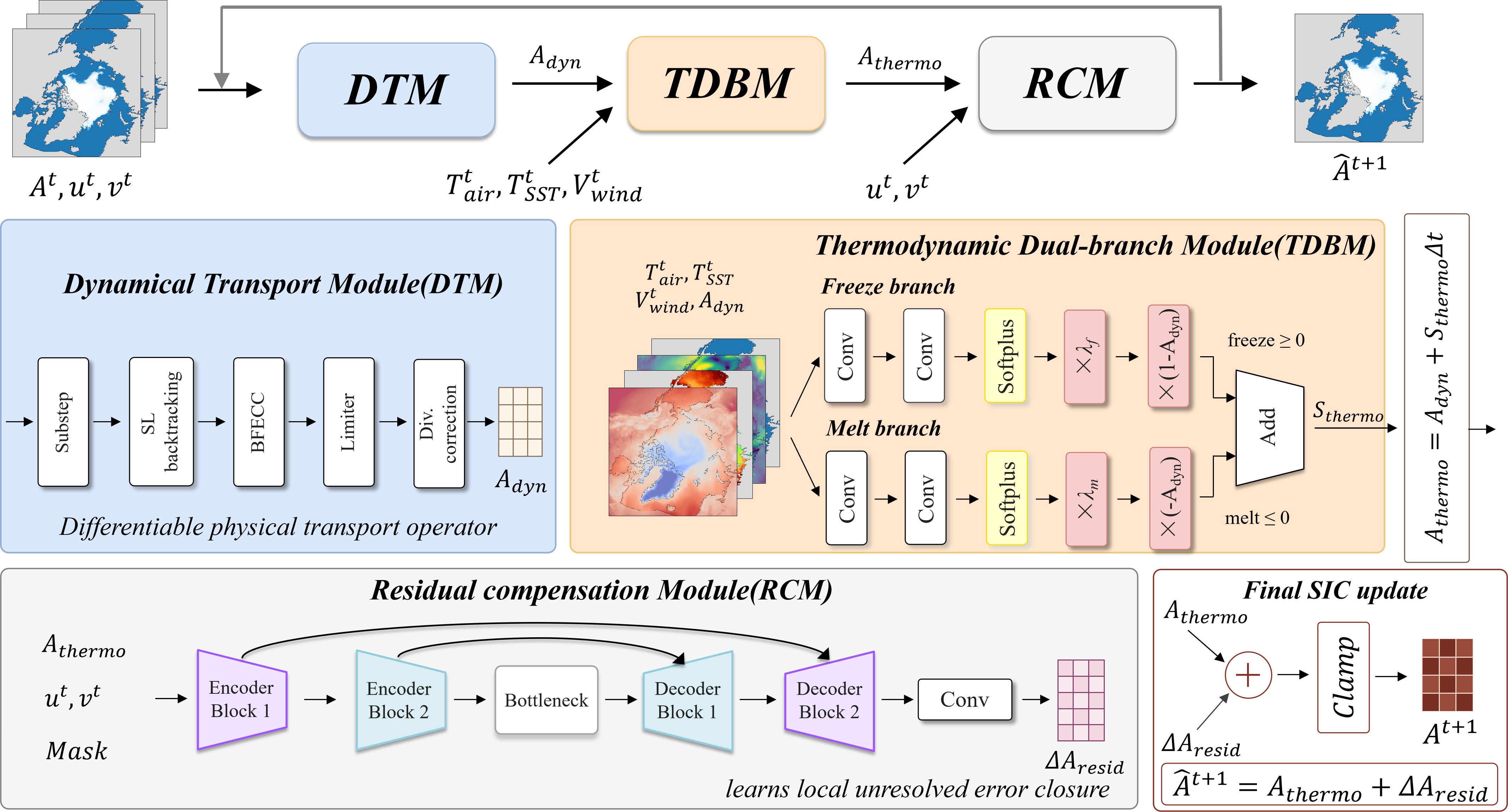}

\caption{PIHIM network overview. (a) The conceptual structure links the SIC continuity-equation decomposition to movement, freeze/melt processes, and local unresolved changes. (b) The implementation-level architecture presents the corresponding module inputs, outputs, and autoregressive rollout for multi-day prediction.}

\end{figure*}

\subsection{Physics-based Numerical Models and Machine-learning Correction}

Physics-based numerical models remain a central foundation for sea ice prediction because they explicitly represent coupled atmosphere, ocean, and sea ice processes. Seasonal systems such as SEAS5 provide valuable numerical references \cite{ref19}, and multimodel comparisons indicate pan-Arctic skill while remaining sensitive to regional variability, initialization errors, extreme years, and model biases \cite{ref15}. Machine learning has therefore been applied to correct or augment numerical systems, including seasonal sea ice prediction from Earth system models \cite{ref33} and online global ice-ocean or coupled prediction systems \cite{ref34,ref35}. These studies demonstrate the value of combining physical models with machine learning, but most remain embedded in full numerical systems or focus on bias correction and system-level forecast performance, rather than lightweight, differentiable, process-decomposed modeling of daily SIC evolution.

\subsection{Data-driven SIC Prediction and Sea Ice Evolution Simulation}

Deep learning is increasingly applied in sea ice remote sensing, SIC prediction, and sea ice state simulation, including automated sea ice mapping and input selection \cite{ref66}. For prediction-oriented tasks, IceNet produces probabilistic seasonal forecasts \cite{ref16}, and IceMamba integrates attention mechanisms into a state space architecture for seasonal pan-Arctic SIC forecasting \cite{ref70}. For diagnostic attribution rather than forecasting, deep neural networks have been used to reconstruct Arctic sea ice extent variability from observational daily SST anomaly fields across ocean basins and identify influential regions through explainable AI \cite{ref71}. Other studies have explored lightweight U-Net models for short-range daily prediction \cite{ref38}, attention-based SICNet for pan-Arctic SIC prediction \cite{ref17}, and IceFormer with multiscale Transformer representations for subseasonal prediction \cite{ref18}. Related ocean studies have used deep learning for long-lead 3D temperature and remote-sensing reflectance prediction \cite{ref68,ref69}, while FuXi-Ocean provides 6-hourly, $1/12^{\circ}$ global ocean forecasts and addresses cumulative error amplification in autoregressive forecasting \cite{ref72}. Together, these works indicate that data-driven models can efficiently capture complex spatiotemporal dependencies across remote-sensing retrieval and prediction tasks. However, most prediction and simulation models still formulate SIC or sea ice state evolution as an end-to-end mapping, leaving their internal states without direct correspondence to explicit dynamical or thermodynamic processes. This limits process diagnosis during autoregressive integration, under distributional shifts, and near high-gradient ice-edge regions.

\subsection{Physics-guided and Constrained Deep Learning Models}

Within the broader literature on physics-informed machine learning \cite{ref25}, recent sea ice studies have introduced physics-related inputs, task-specific losses, conservation constraints, and process-oriented learning to enhance the credibility of sea ice machine learning. Examples include normalized integrated ice-edge error (NIIEE) loss for ice-edge shape errors \cite{ref40}, thickness-informed seasonal SIC prediction \cite{ref41}, and physical information-guided multi-source fusion for high-resolution SIC estimation and ice-edge representation \cite{ref67}. Beyond sea ice, physics-aware statistical frameworks have used meteorological variables and learned interaction factors as physical priors for multi-temporal-resolution marine heatwave forecasting \cite{ref73}. In physics-guided deep learning, previous studies have explored several effective strategies for incorporating domain knowledge into neural networks, including numerical-model supervision, physics-aware architecture design, physically constrained forward propagation, and PDE-informed priors \cite{ref74,ref75,ref76,ref77}. More broadly, data-driven methods can learn unresolved-process closures under physical constraints in ocean modeling \cite{ref29}, and mass-conserving sea ice emulators can improve interpretability and cross-climate generalization \cite{ref43}. These studies indicate that physical guidance has become an active direction in sea ice and ocean machine learning. Nevertheless, existing approaches usually introduce such guidance through auxiliary variables, losses, conservation formulations, or embedded correction. Direct attempts to organize SIC evolution into differentiable structures that explicitly reflect dynamical and thermodynamic processes under a unified sea ice continuity framework remain limited.









\section{Methodology}

\subsection{Problem Formulation and PIHIM Overview}

This study models the daily Arctic SIC field, denoted by $A\in[0,1]$, where each grid-cell value represents the ice-covered fraction. Following common SIC and MIZ thresholds, cells with SIC not exceeding 15\% are defined as open water (OW), cells with SIC greater than 15\% and lower than 80\% as the marginal ice zone (MIZ), and cells with SIC no lower than 80\% as pack ice (PI) \cite{ref13,ref14}. Given the current SIC field $A^t$, sea ice drift velocity $\mathbf{v}^t=(u^t,v^t)$, and atmospheric-oceanic forcing variables $\mathbf{X}^t$, the objective is to learn the one-day evolution operator in Eq.~\eqref{eq:one-day-operator}:
\begin{equation}
\hat{A}^{t+1} = \mathcal{G}_{\Theta}(A^t, \mathbf{v}^t, \mathbf{X}^t),
\label{eq:one-day-operator}
\end{equation}
where $\mathcal{G}_{\Theta}$ denotes the hybrid model. Multi-step prediction is performed autoregressively by feeding each prediction into the next step. Reanalysis and forecast forcings correspond to different evaluation settings but do not change the PIHIM model structure.

PIHIM is structured around the SIC continuity equation in Eq.~\eqref{eq:sic-continuity} \cite{ref9,ref10,ref27,ref28}:
\begin{equation}
\frac{\partial A}{\partial t} + \nabla \cdot (A\mathbf{v}) = S_A,
\label{eq:sic-continuity}
\end{equation}
where the conservative transport term describes velocity-driven horizontal SIC redistribution, and $S_A$ denotes the thermodynamic area source-sink term associated with freezing and melting. Based on Eq.~\eqref{eq:sic-continuity}, PIHIM explicitly implements dynamical transport and thermodynamic areal change, while a residual module compensates for remaining discrepancies arising from discretization approximations and unresolved local effects.

As summarized in Fig. 2, the dynamical module computes velocity-driven SIC redistribution, the thermodynamic module estimates freeze/melt areal change from external forcings, and the residual module learns remaining local errors. This differentiable process-decomposed design provides the structural basis for module diagnosis, ablation analysis, and long-term autoregressive evaluation.

\subsection{Physics-guided SIC Evolution Framework}

\subsubsection{Dynamical Transport Module}

The dynamical transport module represents SIC redistribution driven by horizontal sea ice drift. When thermodynamic source-sink terms are omitted, SIC follows the conservative continuity equation in Eq.~\eqref{eq:transport-pde} \cite{ref9,ref27,ref28}:
\begin{equation}
\frac{\partial A}{\partial t}+\nabla\cdot(A\mathbf{v})=0,
\label{eq:transport-pde}
\end{equation}
where $\mathbf{v}=(u,v)$ is the sea ice drift velocity. This process is expressed as the differentiable transport operator in Eq.~\eqref{eq:dynamical-operator}:
\begin{equation}
A_{\mathrm{dyn}}=\mathcal{D}(A^t,\mathbf{v}^t).
\label{eq:dynamical-operator}
\end{equation}
where $A_{\mathrm{dyn}}$ is the transported SIC state. In implementation, $\mathcal{D}$ employs a semi-Lagrangian scheme with temporal substepping to enhance daily integration stability \cite{ref44,ref45}. Given a one-day time step $\Delta t$ and $N_{\mathrm{sub}}$ transport substeps, the substep size is
\[
\delta t=\frac{\Delta t}{N_{\mathrm{sub}}}.
\]
This study sets $N_{\mathrm{sub}}=24$.
At each substep, departure points are backtracked along the velocity field, SIC is interpolated at these points, and a divergence correction is applied. The resulting $A_{\mathrm{dyn}}$ is passed to the thermodynamics-guided source-term module.

\subsubsection{Thermodynamic Dual-branch Module(TDBM)}

Dynamical transport cannot represent SIC areal growth or decay caused by freezing and melting. Rather than solving a full sea ice thermodynamic model, which would require additional variables such as thickness, radiation, albedo, snow cover, and melt ponds, PIHIM adopts a lightweight thermodynamics-guided source-term representation for horizontal SIC areal change.

Within a grid cell, SIC change is approximated as the combination of new ice formation over available open-water area and melting over existing ice-covered area. Two branches estimate potential freezing and melting tendencies from thermodynamic forcings as shown in Eq.~\eqref{eq:thermo-branches}:
\begin{equation}
\begin{aligned}
Q_f &= \mathcal{F}_{\theta_f}(\mathbf{X}_{\mathrm{thermo}}^t),\\
Q_m &= \mathcal{F}_{\theta_m}(\mathbf{X}_{\mathrm{thermo}}^t),
\end{aligned}
\label{eq:thermo-branches}
\end{equation}
where $\mathbf{X}_{\mathrm{thermo}}^t$ comprises near-surface air temperature, sea surface temperature, and scalar wind speed, and $\mathcal{F}_{\theta_f}$ and $\mathcal{F}_{\theta_m}$ are the freezing and melting branches, respectively. Their structures are provided in Supplementary Section S7.2. Non-negative transformations enforce the signs of the two tendencies in Eq.~\eqref{eq:thermo-signs}:
\begin{equation}
\begin{aligned}
E_f &= \mathrm{softplus}(Q_f),\\
E_m &= \mathrm{softplus}(Q_m).
\end{aligned}
\label{eq:thermo-signs}
\end{equation}
The transported SIC state weights the available area for each process: freezing scales with $1-A_{\mathrm{dyn}}$, and melting scales with $A_{\mathrm{dyn}}$. The thermodynamic source term is therefore defined by Eq.~\eqref{eq:thermo-source}:
\begin{equation}
S_{\mathrm{thermo}}
=
(1-A_{\mathrm{dyn}})\lambda_f E_f
-
A_{\mathrm{dyn}}\lambda_m E_m,
\label{eq:thermo-source}
\end{equation}
where $\lambda_f$ and $\lambda_m$ are learnable scale factors. In implementation, the thermodynamic source term is evaluated only over the valid thermodynamic active region and is set to zero outside this region; the corresponding masking settings are provided in Supplementary Section S7.2. Thus, the module provides a sign-constrained and area-weighted decomposition of thermodynamic areal change without resolving full thermodynamic physics.

\subsubsection{Residual Compensation and State Update}

After dynamical transport and the thermodynamic source-term update, PIHIM forms a physics-guided intermediate state through the explicit time integration in Eq.~\eqref{eq:thermo-state}:
\begin{equation}
A_{\mathrm{thermo}}
=
A_{\mathrm{dyn}}
+
S_{\mathrm{thermo}}\Delta t.
\label{eq:thermo-state}
\end{equation}
Here, $S_{\mathrm{thermo}}\Delta t$ is the one-day SIC increment induced by thermodynamic areal change. Because this intermediate state cannot fully represent reference SIC evolution, owing to unresolved deformation, ice-edge fragmentation, forcing-related mismatch, interpolation error, and discretization error, a residual network estimates the remaining increment in Eq.~\eqref{eq:residual-increment}:
\begin{equation}
\Delta A_{\mathrm{resid}}
=
\mathcal{R}_{\theta}(A_{\mathrm{thermo}},\mathbf{v}^t).
\label{eq:residual-increment}
\end{equation}
The valid ocean mask is applied in implementation. The velocity input allows the residual module to compensate for numerical errors associated with the discretized dynamical-transport step. The residual module is not supplied with raw thermodynamic forcing variables; thermodynamic information enters the residual pathway only indirectly through $A_{\mathrm{thermo}}$.

The final one-step prediction in Eq.~\eqref{eq:final-update} combines the physics-guided state and residual correction:
\begin{equation}
\hat{A}^{t+1}
=
A_{\mathrm{thermo}}
+
\Delta A_{\mathrm{resid}}.
\label{eq:final-update}
\end{equation}
In implementation, \(A_{\mathrm{dyn}}\), \(A_{\mathrm{thermo}}\), and \(\hat{A}^{t+1}\) are clipped to \([0,1]\) and multiplied by the valid ocean mask.

The last layer of the residual network is zero-initialized so that the initial residual is close to zero. This initialization reduces the tendency for the residual module to override the dynamical and thermodynamic baseline at the beginning of training, while allowing it to learn remaining local biases during joint optimization.

Overall, the dynamical module accounts for horizontal transport and ice-edge displacement, the thermodynamic module accounts for freeze/melt areal change, and the residual module compensates for remaining errors. Detailed module architectures, transport settings, and the structural comparison with IceNet are provided in Supplementary Section S7.

\subsection{Training Strategy and Loss Function}

PIHIM combines a parameter-free dynamical transport operator with trainable thermodynamic and residual modules. Because the residual network has substantial representational capacity, direct end-to-end optimization may cause the residual pathway to dominate error gradients before the thermodynamics-guided source-term module learns stable freeze/melt tendencies. PIHIM is therefore trained with staged optimization, autoregressive rollout, and a regional area-bias constraint.

\subsubsection{Training Strategy}

Training contains a warm-up stage and a joint fine-tuning stage. During warm-up, the residual module is frozen and only the thermodynamics-guided source-term module is optimized, encouraging the model to learn forcing-driven freezing and melting tendencies from the dynamical-thermodynamic baseline. During joint fine-tuning, the residual module is unfrozen and optimized together with the thermodynamic module, so that it mainly learns remaining errors rather than replacing the process-decomposed components.

To enhance multi-step stability, training employs autoregressive rollout. Starting from $A^t$, the model is integrated for $T$ consecutive steps, and each predicted SIC field is reused as the next input according to Eq.~\eqref{eq:autoregressive-rollout}:
\begin{equation}
\hat{A}^{t+\tau}
=
\mathcal{G}_{\Theta}
(
\hat{A}^{t+\tau-1},
\mathbf{v}^{t+\tau-1},
\mathbf{X}^{t+\tau-1}
),
\quad
\tau=1,\ldots,T.
\label{eq:autoregressive-rollout}
\end{equation}
where $\hat{A}^{t}=A^t$. This study sets $T=14$, exposing the model to accumulated rollout errors during training. Training objectives, auxiliary constraints, hyperparameters, hardware configuration, and convergence diagnostics are provided in Supplementary Sections S6.2--S6.3.

\subsubsection{Loss Function}

SIC prediction requires both pixel-scale accuracy and control of regional areal drift. Pixel-wise mean squared error constrains local errors but may not sufficiently penalize systematic overestimation or underestimation of regional SIC during multi-step integration. PIHIM therefore adopts a composite loss with a pixel-wise term and a regional area-bias term.

For rollout step $\tau$, Eq.~\eqref{eq:step-loss} defines the single-step loss:
\begin{equation}
\mathcal{L}_{\mathrm{step}}^{(\tau)}
=
\mathcal{L}_{\mathrm{pix}}^{(\tau)}
+
\lambda
\mathcal{L}_{\mathrm{bias}}^{(\tau)},
\label{eq:step-loss}
\end{equation}
where $\lambda$ balances the two terms. The pixel-wise loss in Eq.~\eqref{eq:pixel-loss} is
\begin{equation}
\mathcal{L}_{\mathrm{pix}}^{(\tau)}
=
\frac{1}{\sum M}
\sum
\left[
(\hat{A}^{t+\tau}-A^{t+\tau})\odot M
\right]^2,
\label{eq:pixel-loss}
\end{equation}
where $M$ excludes land and invalid grid cells, and $\odot$ denotes element-wise multiplication. The regional area-bias loss in Eq.~\eqref{eq:bias-loss} is
\begin{equation}
\mathcal{L}_{\mathrm{bias}}^{(\tau)}
=
\left(
\frac{\sum(\hat{A}^{t+\tau}\odot M)}{\sum M}
-
\frac{\sum(A^{t+\tau}\odot M)}{\sum M}
\right)^2.
\label{eq:bias-loss}
\end{equation}
This term penalizes the squared difference between predicted and reference mean SIC over the valid ocean region. The main rollout loss in Eq.~\eqref{eq:rollout-loss} is averaged over all rollout steps:
\begin{equation}
\mathcal{L}
=
\frac{1}{T}
\sum_{\tau=1}^{T}
\mathcal{L}_{\mathrm{step}}^{(\tau)}.
\label{eq:rollout-loss}
\end{equation}
This main rollout loss constrains step-wise spatial accuracy and multi-step regional stability; auxiliary training terms are described in Supplementary Section S6.2.

\section{Experiments}

This section evaluates PIHIM in terms of short-range accuracy, autoregressive stability, process interpretation, and forecast-forced prediction. Subsequent experiments follow the two evaluation settings defined in Section 4.1: reanalysis-forced simulation examines SIC evolution under reanalysis forcing, and forecast-forced prediction assesses D1-D9 short-range performance under forecast-forced conditions \cite{ref46,ref47,ref48,ref49,ref51}.
PIHIM is compared with a persistence baseline and representative data-driven spatiotemporal prediction models, including IceNet \cite{ref16}, SimVP \cite{ref52}, FCNet \cite{ref53}, PredRNNv2 \cite{ref54}, VMRNN \cite{ref55}, SwinLSTM \cite{ref56}, and PredRNN++ \cite{ref57}.

\subsection{Experimental Design}

The experiments distinguish reanalysis-forced simulation from forecast-forced prediction to avoid interpreting prescribed-forcing simulation as independent prediction performance. In reanalysis-forced simulation, ERA5/GLORYS reanalysis forcings, including sea ice drift, are supplied at each valid time. In forecast-forced prediction, each case starts from the initial SIC and analyzed forcing state available at initialization; subsequent D1-D9 rollout is driven by AIFS/RTOFS forecast forcings (RTOFS product information: \url{https://polar.ncep.noaa.gov/global/about/}), while future ERA5, GLORYS, and OSI SAF fields are reserved only for verification.

Within each setting, all trainable baselines are evaluated using the same data splits, input variables, evaluation windows, and verification references as PIHIM. Table 1 summarizes the main settings, and detailed protocols, data roles, metric definitions, information-availability constraints, preprocessing, and vector-rotation details are provided in Supplementary Sections S1 and S6.1.

The metrics include mean absolute error (MAE), root-mean-square error (RMSE), anomaly correlation coefficient (ACC), Integrated Ice Edge Error (IIEE), sea ice area (SIA), and sea ice extent (SIE); area and extent diagnostics are reported in \(10^6\ \mathrm{km^2}\).





















\begin{table*}[t]
\footnotesize

\pihimtablecaption[tbl:settings]{Summary of experimental settings, data, metrics, and baselines}

{
\centering
\setlength{\tabcolsep}{3pt}
\renewcommand{\arraystretch}{1.08}

\begin{tabularx}{\textwidth}{@{}p{0.18\textwidth}p{0.40\textwidth}Y@{}}
\toprule
Category & Setting & Role \\
\midrule
Evaluation setting & reanalysis-forced simulation & Setting defined in Section 4.1 for assessing SIC evolution under reanalysis forcing \\
Evaluation setting & forecast-forced prediction & Setting defined in Section 4.1 for assessing D1-D9 short-range prediction under forecast-forced conditions \\
Reanalysis/reference & ERA5 \cite{ref46}, GLORYS \cite{ref47} & ERA5 provides atmospheric forcings; GLORYS provides SST, SIC, and sea ice velocity \\
Forecast forcings & AIFS \cite{ref48}, RTOFS \cite{ref49} & AIFS provides air temperature and wind speed, RTOFS provides SST, and the sea ice velocity channel is set to zero \\
Observational verification & OSI SAF \cite{ref51} & Satellite SIC for assessing whether the results depend on the GLORYS reference \\
Numerical reference & SEAS5 \cite{ref19} & Serves as a contextual numerical reference rather than a task-equivalent baseline \\
Metrics & MAE, RMSE, Bias, ACC, IIEE, SIA/SIE bias, trend error & Evaluate grid errors, spatial structure, ice-edge error, area/extent drift, and long-term trend \\
Model baselines & Persistence, IceNet, SimVP, FCNet, PredRNNv2, VMRNN, SwinLSTM, PredRNN++ & Cover persistence and data-driven spatiotemporal prediction baselines \\
\bottomrule
\end{tabularx}
}

\end{table*}

\subsection{Reanalysis-forced Simulation}

This section adopts the reanalysis-forced simulation setting defined in Section 4.1 to evaluate PIHIM. The 14-day benchmark assesses short-range SIC evolution accuracy, while the 90-day, calendar-year, and multi-year integrations examine autoregressive rollout stability. Supplementary initial-condition perturbation and extreme-year withholding diagnostics are provided in Supplementary Sections S2.4 and S2.6.

\subsubsection{Fourteen-day Benchmark}

The 14-day benchmark compares PIHIM with Persistence and selected spatiotemporal prediction baselines under reanalysis-forced simulation, with the results reported in Table 2.

\begin{table}[t]

\small

\centering

\pihimtablecaption{Fourteen-day reanalysis-forced simulation benchmark}

\begin{adjustbox}{max width=\linewidth}

\begin{tabular}{lccccc}

\toprule

Model & MAE & RMSE & Bias & ACC & IIEE \\

\midrule

Persistence & 0.0389 & 0.1094 & 0.0009 & 0.6999 & 0.6202 \\

PredRNN++ & 0.0297 & 0.0799 & 0.0028 & 0.8256 & 0.4227 \\

VMRNN & 0.0294 & 0.0714 & 0.0024 & 0.8594 & 0.3536 \\

SwinLSTM & 0.0294 & 0.0712 & 0.0023 & 0.8604 & 0.3476 \\

PredRNNv2 & 0.0277 & 0.0703 & 0.0008 & 0.8627 & 0.3673 \\

FCNet & 0.0265 & 0.0714 & 0.0005 & 0.8600 & 0.3431 \\

SimVP & 0.0247 & 0.0655 & 0.0009 & 0.8809 & 0.2795 \\

IceNet & 0.0238 & 0.0621 & 0.0020 & 0.8916 & 0.2680 \\

PIHIM & \textbf{0.0229} & \textbf{0.0592} & \textbf{0.0004} & \textbf{0.9057} & \textbf{0.2437} \\

\bottomrule

\end{tabular}

\end{adjustbox}

\end{table}

Among the evaluated models, PIHIM attains the lowest MAE, RMSE, and IIEE and the highest ACC. Relative to Persistence, it reduces MAE and IIEE by about 41\% and 61\%, respectively, indicating that it does not simply preserve the initial SIC field. IceNet is the closest baseline, but PIHIM still attains lower MAE and IIEE and higher ACC. Calendar-aligned lead-day heatmaps for the five metrics are provided in Supplementary Section S2.5 to indicate that these averaged results are not dominated by isolated seasonal periods.

\subsubsection{Seasonal-scale and Long-term Stability}

Longer integrations assess the autoregressive stability of PIHIM beyond the 14-day benchmark. SIA is computed as the SIC-weighted area over the valid ocean region, whereas SIE is computed as the area of grid cells with SIC at least 15\% (NSIDC terminology: \url{https://nsidc.org/learn/ask-scientist/what-difference-between-sea-ice-area-and-extent}); both are reported in \(10^6\ \mathrm{km^2}\). The 90-day rollout evaluates intraseasonal stability, calendar-year integration evaluates full annual freeze-melt cycles, and multi-year continuous integration diagnoses long-term areal drift without annual reset. Extended lead-time curves, annual statistics, and multi-year trend diagnostics are provided in Supplementary Sections S2.1--S2.3.

\paragraph{Ninety-day Autoregressive Rollout}

The 90-day rollout comprises 12 seasonal cases initialized on January 1, April 1, July 1, and October 1 in each year from 2022 to 2024. Table 3 reports endpoint-averaged metrics at Day 30, Day 60, and Day 90.

\begin{table}[t]

\small

\centering

\pihimtablecaption{Endpoint diagnostics for 90-day reanalysis-forced simulation}

\begin{adjustbox}{max width=\linewidth}

\begin{tabular}{lccccc}

\toprule

Lead time & MAE & RMSE & ACC & IIEE & SIA bias \\

\midrule

Day 30 & 0.0270 & 0.0679 & 0.8710 & 0.2692 & 0.0157 \\

Day 60 & 0.0272 & 0.0710 & 0.8742 & 0.3038 & 0.0332 \\

Day 90 & 0.0257 & 0.0677 & 0.8808 & 0.2665 & 0.0057 \\

\bottomrule

\end{tabular}

\end{adjustbox}

\end{table}

PIHIM exhibits no apparent error divergence through Day 90. MAE and RMSE remain near 0.026-0.027 and 0.068-0.071, ACC remains above 0.87, and IIEE remains near \(0.27-0.30\times10^6\ \mathrm{km^2}\). The Day-90 metrics should be interpreted as endpoint averages over seasonal cases, not as cumulative averages along a single error sequence.

\paragraph{Calendar-year Integration}

Calendar-year integration starts at the beginning of each test year and continues to year end, evaluating whether PIHIM captures summer melt, autumn-winter freeze-up, and intra-annual SIA/SIE variability. Table 4 reports the corresponding year-wise diagnostic metrics.

\begin{table}[t]

\small

\centering

\pihimtablecaption{Calendar-year integration diagnostics for 2022-2024}

\begin{adjustbox}{max width=\linewidth}

\begin{tabular}{lcccc}

\toprule

Year & MAE & ACC & IIEE & End-of-year SIA bias \\

\midrule

2022 & 0.0276 & 0.8478 & 0.2980 & 0.0252 \\

2023 & 0.0260 & 0.8869 & 0.2709 & 0.0086 \\

2024 & 0.0262 & 0.8917 & 0.2700 & 0.0313 \\

\bottomrule

\end{tabular}

\end{adjustbox}

\end{table}

\begin{figure}

\centering

\includegraphics[width=\linewidth]{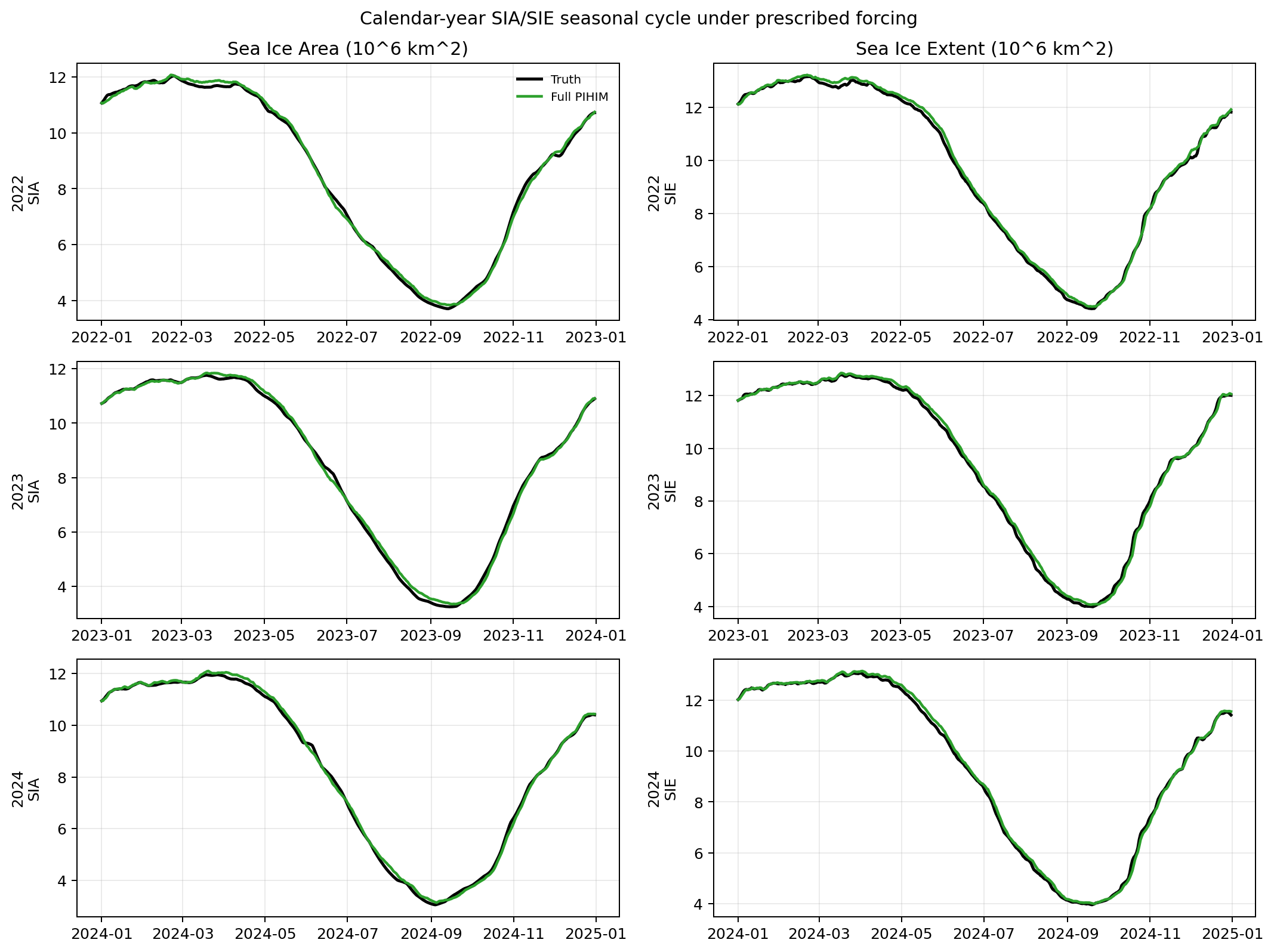}

\caption{Seasonal cycles of SIA/SIE in calendar-year integrations from 2022 to 2024}

\end{figure}

As shown in Fig. 3, the SIA/SIE curves reproduce the main annual phases, indicating that PIHIM maintains stable areal evolution over complete annual cycles. Across 2022-2024, MAE remains around 0.026-0.028, and IIEE remains around \(0.27-0.30\times10^6\ \mathrm{km^2}\). The lower ACC and higher IIEE in 2022 also indicate that long-term performance remains affected by interannual ice-edge conditions and external forcing states.

\paragraph{Multi-year Continuous Integration}

Multi-year continuous integration evaluates stability without annual reset over the 2022-2024 test period. Table 5 reports linear SIA trends for the reference and PIHIM simulation.

\begin{table}[t]

\small

\centering

\pihimtablecaption{Trend-stability diagnostics for multi-year continuous integration}

\begin{adjustbox}{max width=\linewidth}

\begin{tabular}{lccc}

\toprule

Continuous integration period & Truth SIA trend & PIHIM SIA trend & SIA trend error \\

\midrule

2022-2024 & -0.1318 & -0.1193 & 0.0125 \\

\bottomrule

\end{tabular}

\end{adjustbox}

\end{table}

\begin{figure}

\centering

\includegraphics[width=\linewidth]{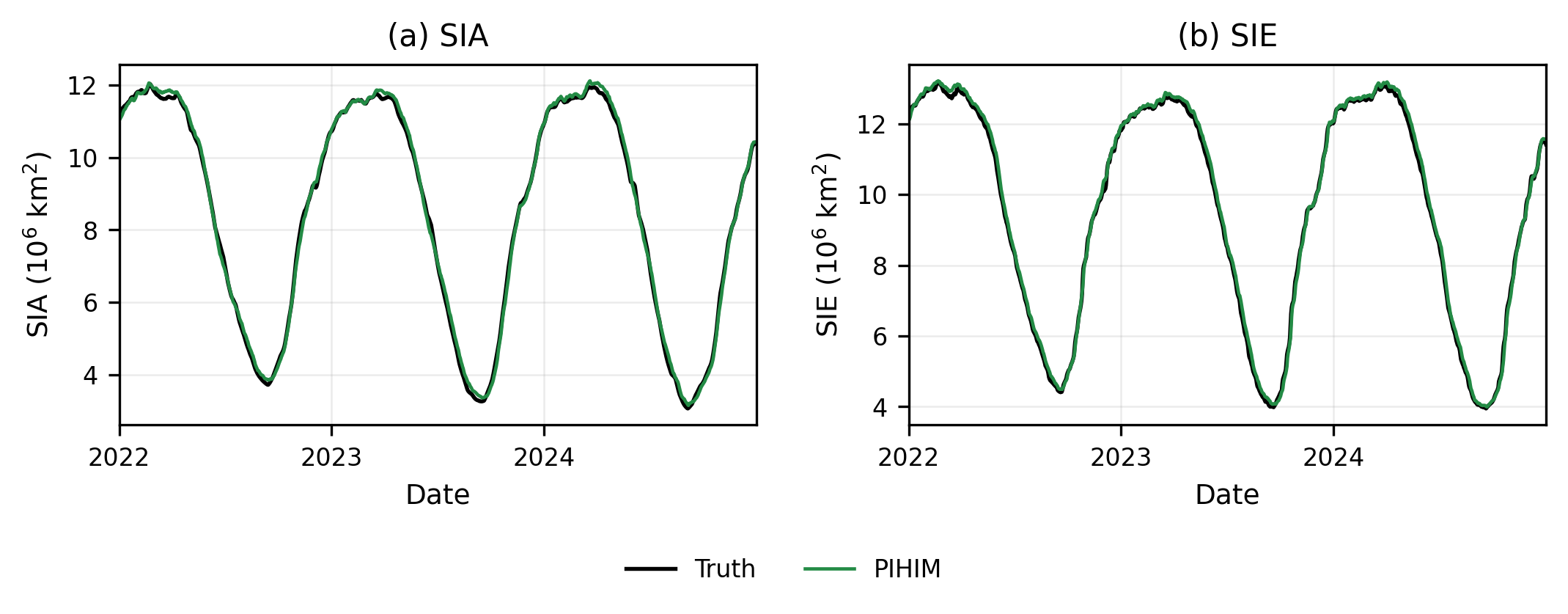}

\caption{SIA/SIE trajectories in multi-year continuous integration over the 2022-2024 test period}

\end{figure}

Fig. 4 shows that PIHIM reproduces the reference trend direction, with a SIA trend error of \(0.0125\times10^6\ \mathrm{km^2}\ \mathrm{yr}^{-1}\). This indicates limited areal drift under prescribed reanalysis forcings. Because the diagnostic covers only the 2022-2024 test period and remains forcing-dependent, it should be interpreted as a stability diagnostic rather than evidence of climate-scale trend prediction capability.

\subsection{Mechanistic Interpretation}

This section interprets PIHIM behavior through component ablation, seasonal-regional error attribution, and thermodynamic source-term diagnostics. The central questions are whether the model gain arises from coupled process modules, where the main errors occur, and whether the thermodynamic outputs exhibit reasonable seasonal phase and perturbation responses. Complete diagnostics are provided in Supplementary Sections S3 and S8.

\subsubsection{Component Contribution}

Table 6 reports inference-time diagnostic ablations on the same trained PIHIM. Specific modules are disabled or retained to examine their immediate contributions; the variants are not treated as independently retrained architectures for model selection.

\begin{table}[t]

\small

\centering

\pihimtablecaption{Inference-time diagnostic ablation results}

\begin{adjustbox}{max width=\linewidth}

\begin{tabular}{lcccc}

\toprule

Variant & MAE & RMSE & ACC & IIEE \\

\midrule

Dynamic + Thermo & 0.0580 & 0.1398 & 0.5725 & 0.6355 \\

Thermo + compensation & 0.0340 & 0.0832 & 0.8275 & 0.2875 \\

Dynamic + compensation & 0.0290 & 0.0708 & 0.8586 & 0.2750 \\

Full PIHIM & \textbf{0.0229} & \textbf{0.0592} & \textbf{0.9057} & \textbf{0.2437} \\

\bottomrule

\end{tabular}

\end{adjustbox}

\end{table}

The full model attains the lowest errors and the highest ACC among these diagnostic variants. Dynamic + Thermo has higher errors and lower ACC than the full model, indicating that explicit process terms alone do not close all SIC evolution errors. Dynamic + compensation and Thermo + compensation also remain below full PIHIM, indicating that dynamical transport, thermodynamic source terms, and residual compensation are complementary. Retrained structural ablations are provided in Supplementary Section S3.1.

\subsubsection{Seasonal and Regional Error Attribution}

Table 7 decomposes errors by ice regime (OW, MIZ, and pack ice) and season (melting, freezing, and transition).

\begin{table}[t]

\small

\centering

\pihimtablecaption{Seasonal-regional error attribution}

\begin{adjustbox}{max width=\linewidth}

\begin{tabular}{llccc}

\toprule

Region & Season & MAE & RMSE & Bias \\

\midrule

MIZ & transition & 0.1384 & 0.1820 & 0.0167 \\

MIZ & freezing & 0.1302 & 0.1685 & -0.0011 \\

MIZ & melting & 0.1192 & 0.1612 & 0.0113 \\

Pack ice & transition & 0.0265 & 0.0474 & 0.0002 \\

Pack ice & freezing & 0.0227 & 0.0426 & -0.0045 \\

Pack ice & melting & 0.0347 & 0.0615 & -0.0138 \\

Open water & transition & 0.0087 & 0.0500 & 0.0076 \\

Open water & freezing & 0.0062 & 0.0328 & 0.0016 \\

Open water & melting & 0.0056 & 0.0372 & 0.0050 \\

\bottomrule

\end{tabular}

\end{adjustbox}

\end{table}

Errors are concentrated in the MIZ, where MAE reaches 0.1192-0.1384 across seasons. Open-water errors are substantially lower, with MAE near 0.006, and pack ice errors remain relatively low. Thus, the primary modeling challenge does not lie in stable open water or the pack ice interior, but in the rapidly evolving ice-edge zone, where low-concentration ice, contour displacement, and freeze-melt phase changes interact.

\subsubsection{Thermodynamic Source-term Diagnostics}

The thermodynamic module is a sign-constrained dual-branch source term with SIC area-fraction weighting: the freezing branch contributes non-negative increments weighted by open-water fraction, and the melting branch contributes non-positive increments weighted by ice-covered fraction. The Table 8 reports seasonal mean source terms, and Table 9 reports responses to thermodynamic forcing perturbations.

\begin{table}[t]

\small

\centering

\pihimtablecaption{Seasonality of thermodynamic source terms}

\begin{adjustbox}{max width=\linewidth}

\begin{tabular}{lccc}

\toprule

Condition & Mean freeze source & Mean melt source & Mean combined source \\

\midrule

Melting season & 0.00265 & -0.00320 & -0.00050 \\

Freezing season & 0.01828 & -0.00125 & 0.01673 \\

Transition season & 0.00905 & -0.00113 & 0.00786 \\

\bottomrule

\end{tabular}

\end{adjustbox}

\end{table}

\begin{table}[t]

\small

\centering

\pihimtablecaption{Response of thermodynamic source terms to external thermodynamic forcing perturbations}

\begin{adjustbox}{max width=\linewidth}

\begin{tabular}{lc}

\toprule

Perturbation experiment & Mean combined source response \\

\midrule

Tair +1 \(^{\circ}\mathrm{C}\) & -0.00091 \\

SST +0.5 \(^{\circ}\mathrm{C}\) & -0.00490 \\

\bottomrule

\end{tabular}

\end{adjustbox}

\end{table}

The net source is negative in the melting season, positive in the freezing season, and intermediate during the transition season. Warming perturbations in Tair and SST both shift the net source toward melting, with a stronger response to SST. These results indicate that the thermodynamic source term has the expected seasonal phase and perturbation direction. Supplementary diagnostics further show that the weak global-mean melt signal mainly reflects post-transport correction demand and dynamical-state effects, rather than insufficient melting-response capacity of the dual-branch structure.

\subsection{Forecast-forced Prediction}

Following the forecast-forced prediction setting in Section 4.1 and the information-availability constraints in Supplementary Section S1.3, forecast sea ice velocity is excluded because short-range drift is sensitive to wind, ocean currents, internal ice mechanics, and data assimilation differences, with reported error growth in drift/deformation forecasts \cite{ref58,ref59,ref60}. Additional diagnostics based on RTOFS forecast sea ice velocity forcing are provided in Supplementary Section S9. Incorporating reliable forecast sea ice velocity and recalibrating the dynamical module are left for future work.

Two test sets are reported: a 99-case set initialized from 2024-04-01 to 2026-04-15 for overall forecast-forced evaluation, and a SEAS5-intersection 25-case set for same-date contextual comparison. GLORYS and OSI SAF serve as verification references; complete test-set definitions and supplementary results are provided in Supplementary Section S4.

\subsubsection{99-case Forecast-forced Prediction Results}

Table 10 and Table 11 report the D1-D9 mean results for the 99-case set under GLORYS and OSI SAF verification, respectively.

PIHIM obtains the lowest MAE, RMSE, and IIEE and the highest ACC under both verification references. Although all models show larger errors under OSI SAF than under GLORYS, PIHIM retains a consistent advantage, suggesting that its performance is not specific to the GLORYS reference.

\begin{table}[t]

\small

\centering

\pihimtablecaption{Comparison of 99-case forecast-forced prediction under GLORYS verification}

\begin{adjustbox}{max width=\linewidth}

\begin{tabular}{lccccc}

\toprule

Model & MAE & RMSE & Bias & ACC & IIEE \\

\midrule

IceNet & 0.0477 & 0.1112 & -0.0106 & 0.7065 & 0.5967 \\

SimVP & 0.0457 & 0.1077 & 0.0197 & 0.7283 & 0.5799 \\

VMRNN & 0.0399 & 0.0988 & 0.0100 & 0.7576 & 0.5329 \\

SwinLSTM & 0.0386 & 0.0972 & 0.0080 & 0.7633 & 0.5196 \\

FCNet & 0.0385 & 0.0980 & \textbf{0.0030} & 0.7372 & 0.5336 \\

PredRNNv2 & 0.0367 & 0.0942 & 0.0098 & 0.7778 & 0.5354 \\

PredRNN++ & 0.0356 & 0.0958 & 0.0046 & 0.7557 & 0.5288 \\

PIHIM & \textbf{0.0321} & \textbf{0.0912} & 0.0082 & \textbf{0.7839} & \textbf{0.4711} \\

\bottomrule

\end{tabular}

\end{adjustbox}

\end{table}

\begin{table}[t]

\small

\centering

\pihimtablecaption{Comparison of 99-case forecast-forced prediction under OSI SAF verification}

\begin{adjustbox}{max width=\linewidth}

\begin{tabular}{lccccc}

\toprule

Model & MAE & RMSE & Bias & ACC & IIEE \\

\midrule

SimVP & 0.0613 & 0.1437 & 0.0227 & 0.7076 & 0.9656 \\

VMRNN & 0.0594 & 0.1487 & 0.0115 & 0.6843 & 1.0365 \\

SwinLSTM & 0.0591 & 0.1475 & 0.0106 & 0.6887 & 1.0267 \\

FCNet & 0.0586 & 0.1441 & 0.0080 & 0.6879 & 0.9774 \\

PredRNN++ & 0.0572 & 0.1489 & 0.0094 & 0.6707 & 1.0054 \\

IceNet & 0.0554 & 0.1249 & -0.0167 & 0.7939 & 0.9106 \\

PredRNNv2 & 0.0545 & 0.1386 & 0.0113 & 0.7212 & 0.9579 \\

PIHIM & \textbf{0.0448} & \textbf{0.1157} & \textbf{-0.0004} & \textbf{0.8071} & \textbf{0.8473} \\

\bottomrule

\end{tabular}

\end{adjustbox}

\end{table}


\subsubsection{SEAS5-intersection 25-case Contextual Comparison}

The 25-case comparison uses initialization dates common to SEAS5 and PIHIM. SEAS5-EnsMean is the 51-member mean, while SEAS5-BestMember and SEAS5-WorstMember characterize the ensemble range rather than task-equivalent baselines.

\begin{table}[t]

\small

\centering

\pihimtablecaption{SEAS5-intersection 25-case results under GLORYS verification}

\begin{adjustbox}{max width=\linewidth}

\begin{tabular}{lccccc}

\toprule

Model & MAE & RMSE & Bias & ACC & IIEE \\

\midrule

SEAS5-WorstMember & 0.0614 & 0.1345 & -0.0163 & 0.6035 & 0.8566 \\

SEAS5-EnsMean & 0.0597 & 0.1310 & -0.0162 & 0.6148 & 0.8253 \\

SEAS5-BestMember & 0.0597 & 0.1315 & -0.0159 & 0.6169 & 0.8208 \\

IceNet & 0.0479 & 0.1110 & -0.0107 & 0.7073 & 0.5851 \\

SimVP & 0.0463 & 0.1088 & 0.0207 & 0.7286 & 0.5756 \\

VMRNN & 0.0399 & 0.0984 & 0.0106 & 0.7619 & 0.5236 \\

FCNet & 0.0394 & 0.0990 & \textbf{0.0039} & 0.7357 & 0.5303 \\

SwinLSTM & 0.0385 & 0.0966 & 0.0087 & 0.7700 & 0.5075 \\

PredRNNv2 & 0.0367 & 0.0940 & 0.0107 & 0.7824 & 0.5232 \\

PredRNN++ & 0.0356 & 0.0954 & 0.0046 & 0.7629 & 0.5201 \\

PIHIM & \textbf{0.0319} & \textbf{0.0903} & 0.0088 & \textbf{0.7897} & \textbf{0.4561} \\

\bottomrule

\end{tabular}

\end{adjustbox}

\end{table}

\begin{table}[t]

\small

\centering

\pihimtablecaption{SEAS5-intersection 25-case results under OSI SAF verification}

\begin{adjustbox}{max width=\linewidth}

\begin{tabular}{lccccc}

\toprule

Model & MAE & RMSE & Bias & ACC & IIEE \\

\midrule

SEAS5-WorstMember & 0.0675 & 0.1473 & -0.0029 & 0.6810 & 1.2147 \\

SEAS5-EnsMean & 0.0657 & 0.1441 & -0.0036 & 0.6942 & 1.1638 \\

SEAS5-BestMember & 0.0656 & 0.1448 & -0.0034 & 0.6929 & 1.1644 \\

SimVP & 0.0624 & 0.1457 & 0.0242 & 0.7107 & 0.9996 \\

FCNet & 0.0603 & 0.1463 & 0.0097 & 0.6900 & 1.0144 \\

VMRNN & 0.0600 & 0.1499 & 0.0119 & 0.6910 & 1.0779 \\

SwinLSTM & 0.0598 & 0.1486 & 0.0114 & 0.6955 & 1.0668 \\

PredRNN++ & 0.0582 & 0.1501 & 0.0100 & 0.6783 & 1.0393 \\

IceNet & 0.0563 & 0.1264 & -0.0166 & 0.7965 & 0.9390 \\

PredRNNv2 & 0.0551 & 0.1398 & 0.0123 & 0.7277 & 0.9968 \\

PIHIM & \textbf{0.0453} & \textbf{0.1164} & \textbf{-0.0005} & \textbf{0.8129} & \textbf{0.8745} \\

\bottomrule

\end{tabular}

\end{adjustbox}

\end{table}

\begin{figure}

\centering

\includegraphics[width=\linewidth]{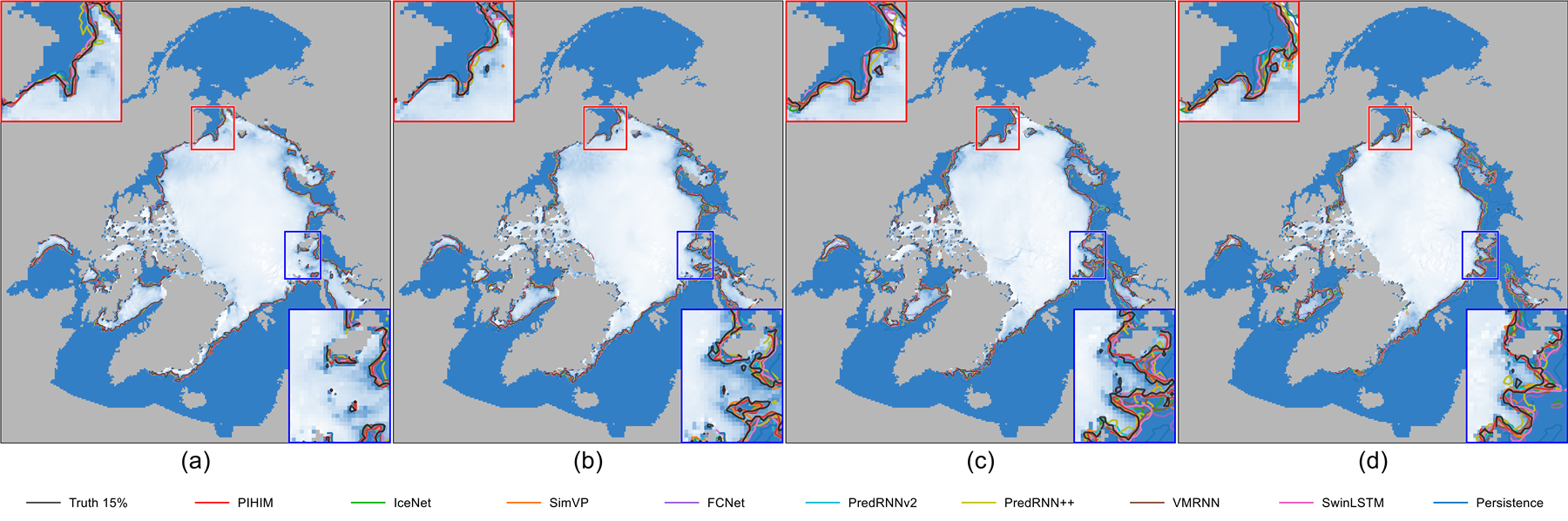}

\caption{Rapid ice-edge retreat case in reanalysis-forced simulation. Panels (a)-(d) correspond to D2, D6, D10, and D14 from the July 2, 2024 initialization.}

\end{figure}

\begin{figure}

\centering

\includegraphics[width=\linewidth]{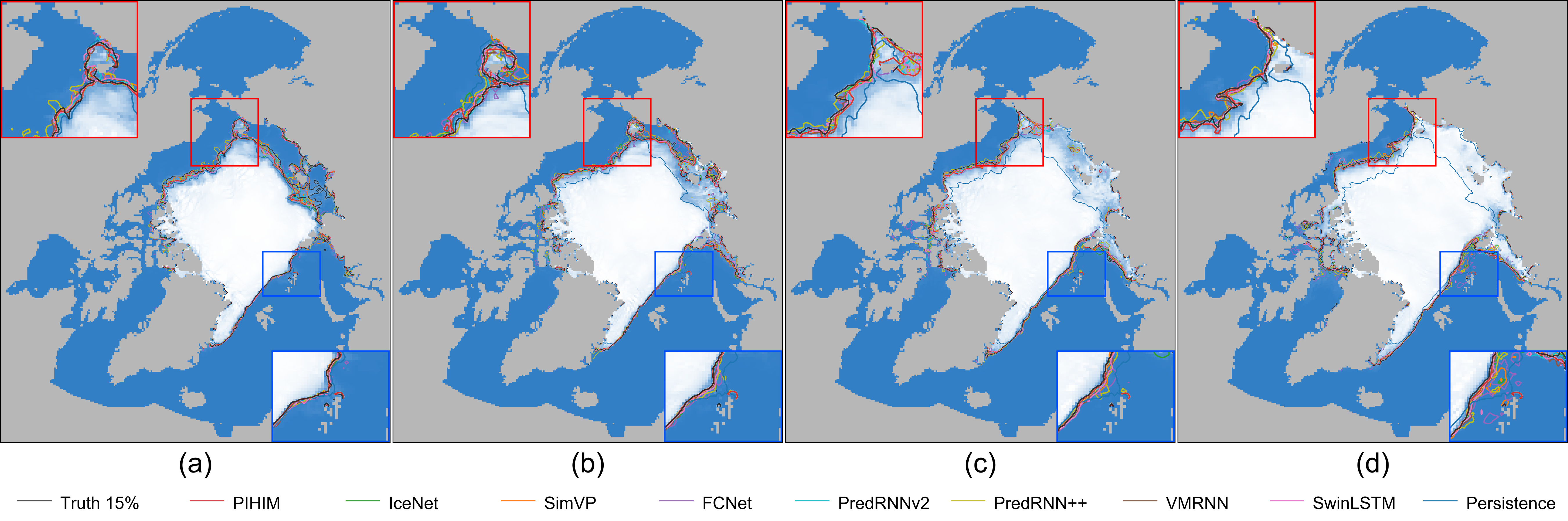}

\caption{Rapid ice-edge expansion case in reanalysis-forced simulation. Panels (a)-(d) correspond to D2, D6, D10, and D14 from the October 16, 2024 initialization.}

\end{figure}

The 25-case results are consistent with the 99-case evaluation. Under both verification references, PIHIM outperforms the selected learning baselines and SEAS5 ensemble summaries, while SEAS5 remains a contextual numerical reference rather than part of the full 99-case comparison.

\subsection{Ice-edge Spatial Case Analysis}

Spatial cases supplement the global statistics by comparing 15\% SIC ice-edge contours and local morphology under rapid retreat/expansion and forecast-forced conditions.

\subsubsection{Rapid Retreat and Expansion Cases in Reanalysis-forced Simulation}

The July 2 and October 16, 2024 cases correspond to the fastest 14-day retreat and expansion diagnostics. Fig. 5 and Fig. 6 show that PIHIM keeps the 15\% contour closer to GLORYS than the baselines under both conditions; complete D1-D14 sequences are provided in Supplementary Section S5.1.











\subsubsection{99-case Forecast-forced Prediction Spatial Case}

Fig. 7 and Fig. 8 present the May 22, 2024 forecast-forced case under GLORYS and OSI SAF verification. PIHIM remains comparatively stable from D3 to D9, consistent with the 99-case statistics; complete sequences are provided in Supplementary Section S5.2.

\begin{figure}

\centering

\includegraphics[width=0.99\linewidth]{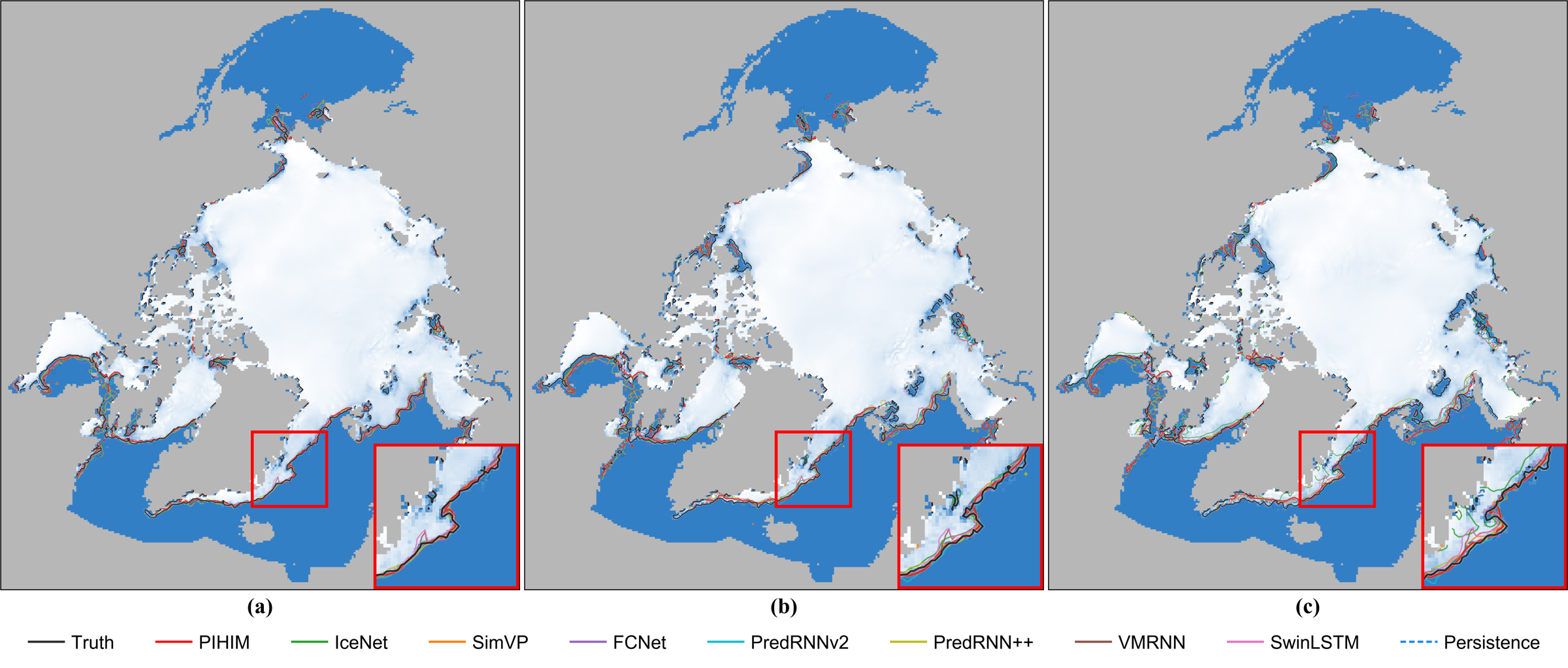}

\caption{GLORYS-verified ice-edge case in 99-case forecast-forced prediction. Panels (a)-(c) correspond to D3, D6, and D9 from the May 22, 2024 initialization.}

\end{figure}

\begin{figure}

\centering

\includegraphics[width=0.99\linewidth]{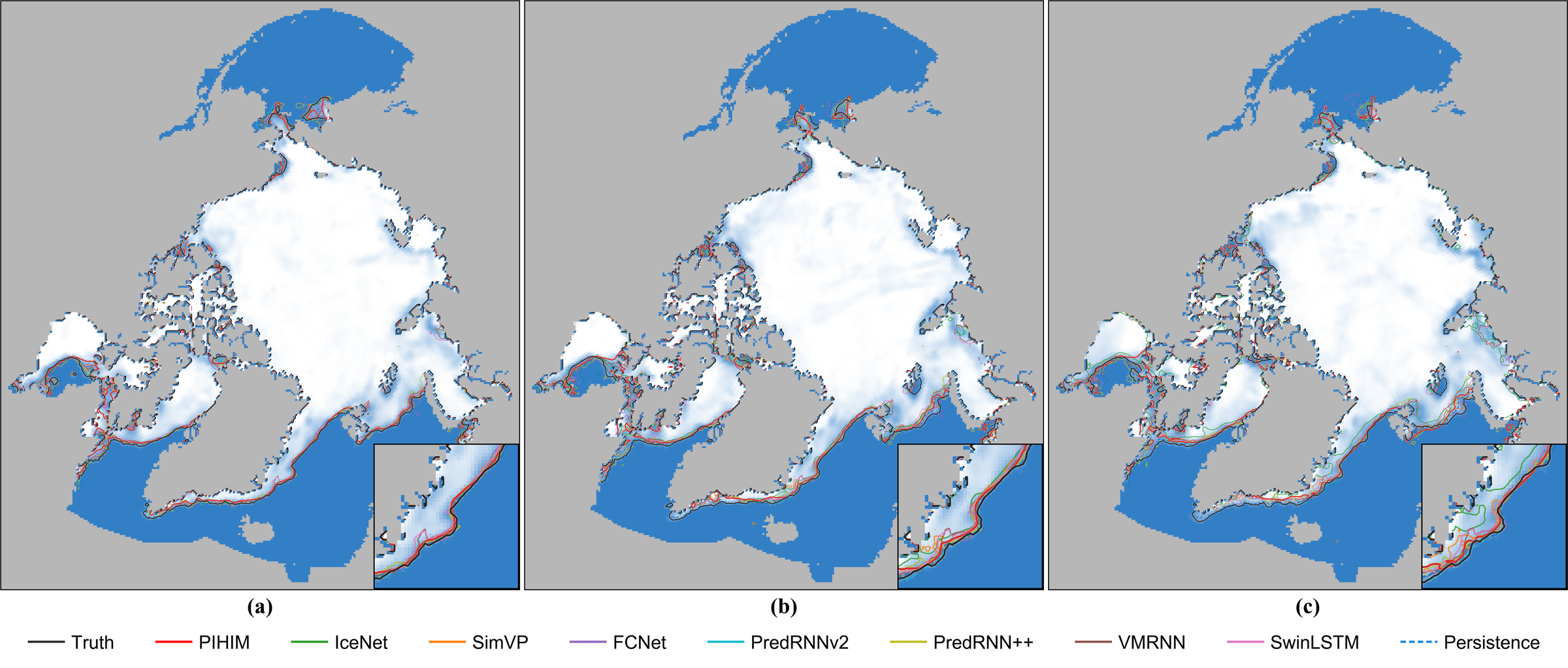}

\caption{OSI SAF-verified ice-edge case in 99-case forecast-forced prediction. Panels (a)-(c) correspond to D3, D6, and D9 from the May 22, 2024 initialization.}

\end{figure}












\subsubsection{Contextual Spatial Comparison on the SEAS5-intersection Samples}

Fig. 9 and Fig. 10 provide contextual spatial comparisons with SEAS5 for the April 1, 2026 initialization on the 25-case set. Under both verification references, PIHIM shows smaller ice-edge displacement than SEAS5 EnsMean, consistent with Table 12 and Table 13; complete D1-D9 sequences are provided in Supplementary Section S5.3.

\begin{figure}
\centering
\includegraphics[width=0.99\linewidth]{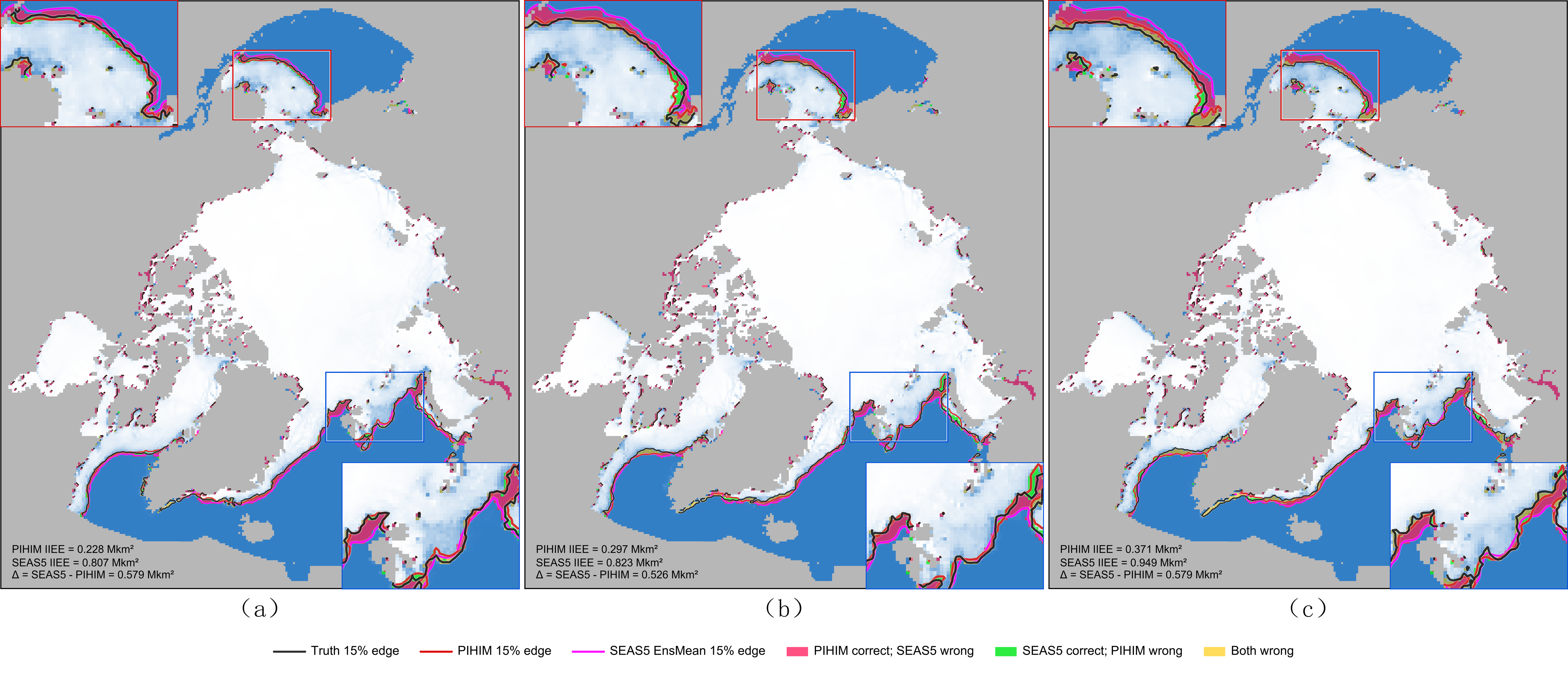}
\caption{GLORYS-verified ice-edge comparison in SEAS5-intersection 25-case forecast-forced prediction. Panels (a)-(c) correspond to D3, D6, and D9 from the April 1, 2026 initialization.}
\end{figure}

\begin{figure}
\centering
\includegraphics[width=0.99\linewidth]{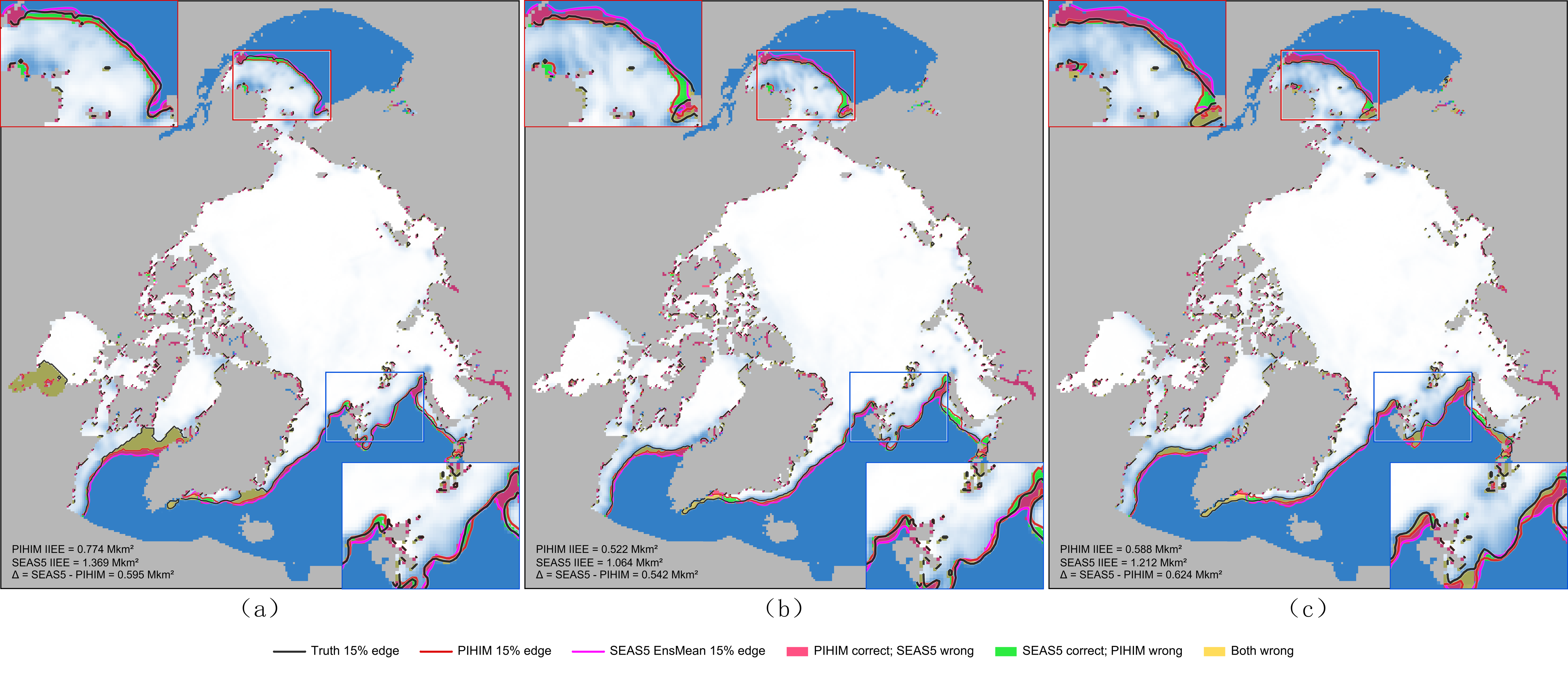}
\caption{OSI SAF-verified ice-edge comparison in SEAS5-intersection 25-case forecast-forced prediction. Panels (a)-(c) correspond to D3, D6, and D9 from the April 1, 2026 initialization.}
\end{figure}

\section{Conclusion}
This study proposed PIHIM for daily Arctic SIC evolution modeling and short-range prediction. Guided by the sea ice continuity equation, PIHIM decomposes SIC evolution into dynamical transport, thermodynamic areal change, and residual compensation, providing a process-decomposed alternative to purely end-to-end SIC sequence models. Experiments show that PIHIM improves ice-edge preservation, error-growth control, and areal stability in reanalysis-forced simulation, while retaining consistent short-range advantages under GLORYS and OSI SAF verification in forecast-forced prediction. Module diagnostics further indicate complementary roles of dynamical transport, thermodynamic source-term representation, and residual compensation in spatial redistribution, freeze/melt areal change, and local error closure.

PIHIM remains a lightweight data-driven model rather than a full numerical sea ice model. Its thermodynamic module does not explicitly resolve sea ice thickness, snow cover, melt ponds, albedo, or fine-scale energy-budget processes, and forecast-forced prediction still depends on external atmospheric and oceanic forecast quality. Future work will incorporate more reliable thickness, energy-budget, and sea ice motion information, refine thermodynamic source-term representation and forecast-forcing error attribution, and explore integration with data assimilation and broader sea ice applications.

\section*{Conflict of Interest}
None of the authors have a conflict of interest to disclose.

\end{document}